\pdfoutput=1

\newif\ifdouble

\doubletrue

\ifdouble
  \documentclass[final,5p,times,twocolumn]{elsarticle}
\else
  \documentclass[preprint,review,12pt]{elsarticle}
\fi

\newcommand{\rv}[1]{}

\usepackage{flushend}

\usepackage{url}

\usepackage[shortlabels]{enumitem}

\usepackage{color, soul}
\definecolor{lightblue}{rgb}{.40,.90,1}
\sethlcolor{lightblue}

\usepackage{color}

\usepackage[cmex10]{amsmath}
\usepackage{amssymb}
\usepackage{nicefrac}

\usepackage[tight,footnotesize]{subfigure}

\usepackage{multirow}

\usepackage{blindtext}

\usepackage{balance}

\usepackage{mathtools}

\DeclarePairedDelimiter\floor{\lfloor}{\rfloor}

\usepackage{amssymb}

\usepackage{booktabs}
\usepackage{multirow}
\usepackage[table,xcdraw]{xcolor}

\graphicspath{{figures/}}

\journal{Elsevier Journal of Signal Processing: Image Communication}

\begin{document}

\begin{frontmatter}



\title{Deep convolutional neural networks for pedestrian detection}


\author[]{D. Tom\`{e}\corref{cor2}}
\ead{denis.tome@mail.polimi.it}

\author[]{F. Monti\corref{cor2}}
\ead{federico2.monti@mail.polimi.it}
 
\author[]{L. Baroffio\corref{cor1}}
\ead{luca.baroffio@polimi.it}

\author[]{L. Bondi}
\ead{luca.bondi@polimi.it}
 
\author[]{M. Tagliasacchi}
\ead{marco.tagliasacchi@polimi.it}

\author[]{S. Tubaro}
\ead{stefano.tubaro@polimi.it}
%
%
%
\address{Dipartimento di Elettronica, Informazione e Bioingegneria, Politecnico di Milano, Piazza Leonardo Da Vinci, 32, 20133, Milan, Italy.}
\cortext[cor2]{Equal contribution}
\cortext[cor1]{Corresponding author}

\begin{abstract}
Pedestrian detection is a popular research topic due to its paramount importance for a number of applications, especially in the fields of automotive, surveillance and robotics. Despite the significant improvements, pedestrian detection is still an open challenge that calls for more and more accurate algorithms. In the last few years, deep learning and in particular convolutional neural networks emerged as the state of the art in terms of accuracy for a number of computer vision tasks such as image classification, object detection and segmentation, often outperforming the previous gold standards by a large margin. In this paper, we propose a pedestrian detection system based on deep learning, adapting a general-purpose convolutional network to the task at hand. By thoroughly analyzing and optimizing each step of the detection pipeline we propose an architecture that outperforms traditional methods, achieving a task accuracy close to that of state-of-the-art approaches, while requiring a low computational time. Finally, we tested the system on an NVIDIA Jetson TK1, a 192-core platform that is envisioned to be a forerunner computational brain of future self-driving cars.


\end{abstract}


\end{frontmatter}


\section{Introduction}
\label{sec:intro}

Humans need just few glances to recognize objects and people, identify events and detect possibly dangerous situations. The correct interpretation of different visual stimuli is key for human to accomplish very complex tasks such as driving a vehicle or playing sports. Furthermore, a large number of tasks require the scene to be analyzed in as few as tens of milliseconds, so as to promptly react to such visual stimuli. Artificial intelligence and in particular computer vision algorithms aim at automatically interpreting the visual content of a scene, in the form of a single frame or a sequence of frames, and react accordingly. The detection of human shapes or pedestrians is one of the most challenging problems that computer vision researchers are tackling since at least two decades ago~\cite{Polana}. It is key to a number of high level applications ranging from car safety to advanced surveillance systems. The last decade~\cite{nguyen2016human} has seen significant improvements of pedestrian detection systems in terms of both accuracy and efficiency, fostered by the advent of more and more powerful yet compact hardware. 

Most pedestrian detection algorithms share similar computation pipelines. First, starting from the raw pixel-level image content, they extract higher-level spatial representations or features resorting to arbitrarily complex transformation to be applied pixel-by-pixel or window-by-window. Second, the features for any given spatial window are fed to a classifier that assesses whether such a region depicts a human. Furthermore, a scale-space is typically used in order to detect pedestrians at different scales, that is, distance with respect to the sensing device. In 2003, Viola and Jones~\cite{VJ} propose a pedestrian detection system based on box-shaped filters, that can be applied efficiently resorting to integral images. The features, i.e. the result of the convolution of a window with a given box-shaped filter, are then fed to a classifier based on AdaBoost~\cite{AdaBoost}. Dalal and Triggs refine the process, proposing Histogram Of Gradients (HOG)~\cite{Dalal} as local image features, to be fed to a linear Support Vector Machine aimed at identifying windows containing humans. Such features proved to be quite effective for the task at hand, representing the basis for more complex algorithms. Felzenswalb et al.~\cite{DPM} further improve the detection accuracy by combining the Histogram Of Gradients with a Deformable Part Model. In particular, such approach aims at identifying a human shape as a deformable combination of its parts such as the trunk, the head, etc. Each body part has peculiar characteristics in terms of its appearance and can be effectively recognized resorting to the HOG features and a properly trained classifier. Such a model proved to be more robust with respect to body shape and pose and to partial occlusions. Doll\'ar et al.~\cite{ICF} propose to use features extracted from multiple different channels. Each channel is defined as a linear or non-linear transformation of the input pixel-level representation. Channels can capture different local properties of the image such as corner, edges, intensity, color. 

Besides the improvements in terms of the quality of visual features, great strides have been made in reducing the computational complexity of the task at hand. For instance, the computation of HOG has been significantly accelerated resorting to fast scale-space approximation algorithms, so as to efficiently estimate local gradients at different scales, leading to the Aggregated Channel Features (ACF)~\cite{ACF}. 
To further boost the performance of pedestrian detection systems, ACF combines HOG and channel features, so as to generate rich representations of the visual content~\cite{ACF}. As a further improvement, Nam et al.~\cite{LDCF} observe that ACF exploits a classifier based on boosting that performs orthogonal splits, i.e., splits based on a single feature element. Instead, they propose to linearly combine the different feature channels so as to remove the correlation to the data, being able to perform oblique splits. Such approach leads to the Locally Decorrelated Channel Features (LDCF)~\cite{LDCF}, that improve the performance of the classifier.

Deep neural network are quickly revolutionizing the world of machine learning and artificial intelligence. They are setting new benchmarks for a number of heterogeneous applications in different areas, including image understanding, speech and audio analysis and natural language processing, filling the gap with respect to human performance for several tasks~\cite{Ranzato::DeepFace}. Despite being around since the 1990s~\cite{lecun95convolutional}, they blossomed in the past few years, in part due to the advent of powerful parallel computation architectures and the development of efficient training algorithms. In particular, Convolutional Neural Networks (CNN) represented a revolution for image analysis. They are considered the state of the art for a number of tasks including image classification~\cite{GoogLeNet}, face recognition~\cite{Ranzato::DeepFace} and object detection~\cite{RCNN}.

In the context of pedestrian detection, there has been a surge of interest in Convolutional Neural Network during the last few years, motivated by the successes for similar image analysis tasks. In particular, object detection and pedestrian detection share a very similar pipeline. For both, some candidate regions have to be identified by means of a sliding window approach or more complex region proposal algorithm. Then, considering object detection, each region should be analyzed to check whether it contains an object and, if so, identify the class of such object. Instead, for pedestrian detection each proposal should be analyzed in order to check whether it contains a human shape. For both tasks such last stage of detection can be effectively accomplished resorting to a properly trained classifier. LeCun et al.~\cite{LeCunPedestrian} were the first to use convolutional networks for detecting pedestrians, proposing an unsupervised deep learning approach. The deformable part model proposed by Felzenswalb et al. has been coupled with a stack of generative, stochastic neural networks and, in particular, Restricted Boltzmann Machine~\cite{DBN-Isol}. A deep stack of networks in place of the original features improves the discriminative ability of the system, while preserving all the advantages of the deformable part model, i.e. robustness to pose and partial occlusions. Such model has been further improved in~\cite{JointDeep} where the authors construct a deep network that is able to perform feature extraction, part deformation handling, and occlusion handling. Hosang et al.~\cite{Hosang2015Cvpr} propose to use a supervised deep learning approach, adapting a network devised for image classification, in order to detect pedestrians. Such approach yields good results in terms of detection accuracy, improving the performance of state-of-the-art methods based on handcrafted features such as LDCF.

In the context of deep learning, small details are often crucial to obtain good results in terms of task accuracy. A small difference in the setting of a single parameter may imply a big difference in the overall performance of the system. In this paper, we build upon the work of Hosang et al.~\cite{Hosang2015Cvpr}, completely dissecting and analyzing their pipeline for pedestrian detection. 

The paper proposes several novel contributions: 
\begin{itemize}
	\item we optimize most of the stages of the pedestrian detection pipeline, proposing novel solutions that significantly improve the detection accuracy\footnote{The source code for our method can be found at \url{https://github.com/DenisTome/DeepPed}}; 
	\item we approach state-of-the-art performance in terms of detection accuracy, outperforming both traditional methods based on handcrafted features and deep learning approaches; 
	\item we propose a lightweight version of our algorithm that runs in real-time on modern hardware;
	\item we validate our approach by implementing it on an NVIDIA Jetson TK1, a compact computational platform based on a Graphics Processing Unit that is being adopted as the computational brain of several car prototypes featuring modern safety systems.
\end{itemize}

The rest of the paper is organized as follows: Section~\ref{sec:background} presents the pipeline that we use for detecting pedestrians, thoroughly illustrating each step, whereas Section~\ref{sec:method} reports all the proposed optimizations that improve the performance of the system. Section~\ref{sec:experiments} is devoted to experimental evaluation, whereas conclusions are drawn in Section~\ref{sec:conclusions}.

\section{Background on pedestrian detection and Convolutional Neural Networks}
\label{sec:background}

\subsection{Pedestrian detection pipeline}

\begin{figure*}[t]
	\centering
    \includegraphics[trim=0cm 0.7cm 0.4cm 0cm, clip=true, width=0.75\textwidth]{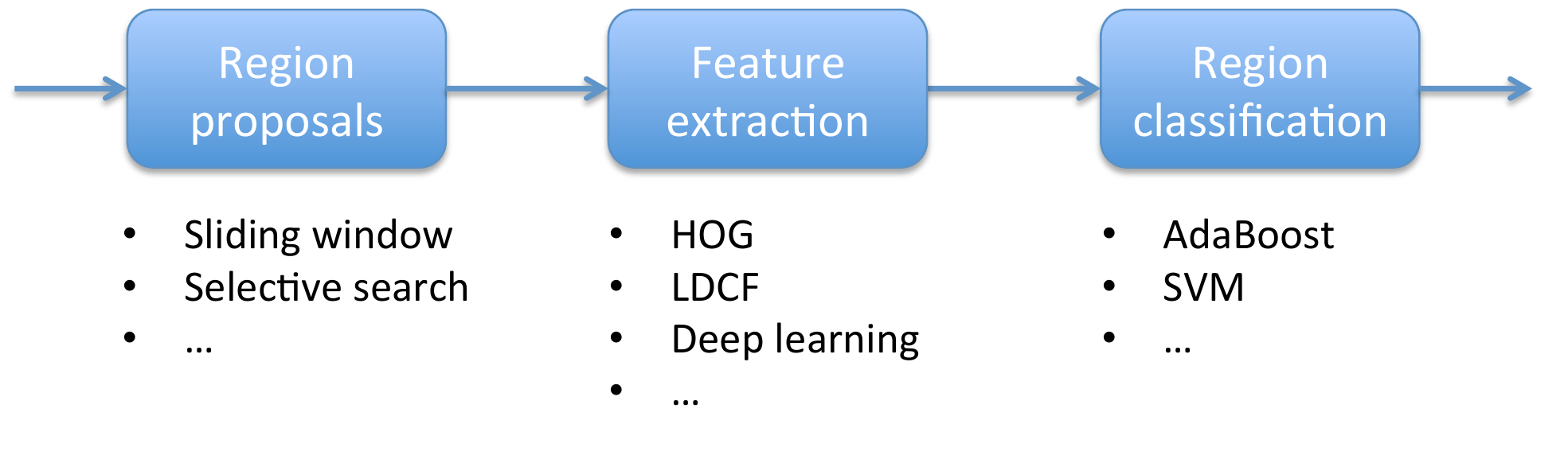}
	\caption{A common pipeline for pedestrian detection.}
	\label{fig:pipeline}
\end{figure*}

During the past two decades, a number of different approaches to pedestrian detection have been proposed and successfully implemented for both commercial and military applications. Despite being very different in the way they process the raw data so as to obtain semantic representations and detect human shapes, they share a similar pipeline for data processing. The input of such pipeline is a raw, pixel-level representation of a scene, whereas the output consists of a set of bounding boxes with different size, each corresponding to a pedestrian that has been identified within the analyzed frame. Such pipeline comprises three main stages: i) region proposal, ii) feature extraction and iii) region classification, as shown in Figure~\ref{fig:pipeline}.

As regards the first stage, i.e. region proposal, the entire frame is analyzed so as to extract candidate regions, i.e. portions of the image that potentially contain a human. The input of such stage is the entire frame, whereas the output is a set of regions, possibly having heterogeneous dimensions and ratios. The sliding window approach is the simplest instance of region proposal algorithms, and can be adapted so as to extract regions at multiple scales and aspect ratios. More complex approaches analyze the visual content to filter out regions that are believed not to contain objects or salient content, so as to reduce the number of candidate regions to be processed at the next stages. Objectness~\cite{alexe2012measuring}, Selective Search~\cite{Uijlings13}, category-independent object proposals~\cite{endres2010category} are instances of such class of algorithms. Such algorithms are general-purpose and thus not tailored to pedestrian detection. Instead, this stage can be substituted with lightweight and efficient algorithms tailored to pedestrian detection, that aim at discarding a high number of negative regions, i.e. the ones not containing a pedestrian, while preserving as many positive regions as possible~\cite{Hosang2015Cvpr}. In this case, the region proposal algorithm acts as a coarse filter that significantly reduces the number of region to be analyzed and thus the computational burden.

As for the feature extraction stage, a number of different methods have been proposed, as mentioned in Section~\ref{sec:intro}. Such methods process the data very differently and exploits disparate visual characteristics, such as local intensity contrast, pooled gradients and multiple non-linear transformations of the input data, in the case of Viola-Jones~\cite{VJ}, Histogram of Gradients~\cite{Dalal} and Integral Channel Features~\cite{ICF}, respectively. The input of such stage is a set of candidate regions, i.e. portions of the input image potentially containing a pedestrian, whereas the output is a feature vector, i.e. a set of real-valued or binary values, for each input region. The feature vector is a compact representation of the visual characteristics of the candidate region.

Finally, the classification stage aims at identifying which regions within the set of candidates correspond to a human shape. The classifier is fed with a feature vector relative to a given region and typically provides a binary label indicating whether such region is positive, i.e. it contains a pedestrian. Early methods such as the one proposed by Viola and Jones~\cite{VJ} exploits AdaBoost, whereas more recent approaches use Support Vector Machines~\cite{Dalal}. In some cases, considering methods based on Convolutional Neural Networks, the classifier is based on hinge or cross-entropy loss functions, resembling support vector machines or logistic regression, respectively, learning both the classifier and the features at once.

\subsection{Background on Convolutional Neural Networks}

Convolutional Neural Networks recorded amazingly good performance in several tasks, including digit recognition, image classification and face recognition. The key idea behind CNNs is to automatically learn a complex model that is able to extract visual features from the pixel-level content, exploiting a sequence of simple operations such as filtering, local contrast normalization, non-linear activation, local pooling. Traditional methods use \emph{handcrafted} features, that is, the feature extraction pipeline is the result of human intuitions and understanding of the raw data. For instance, the Viola-Jones~\cite{VJ} features come from the observation that the shape of a pedestrian is characterized by abrupt changes of pixel intensity in the regions corresponding to the contour of the body. 

\begin{figure*}[t]
	\centering
    \includegraphics[trim=0cm 0cm 0cm 0cm, clip=true, width=1\textwidth]{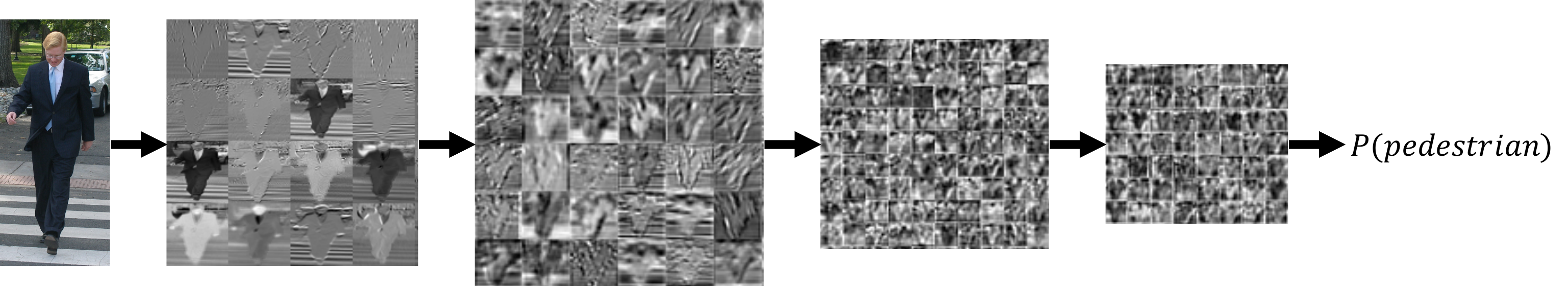}
	\caption{An example of feature maps obtained at each layer of a Convolutional Neural Network. Former layers (left) identify simple structures such as edges and details, whereas next layers identify more complex visual concepts.}
	\label{fig:CNN}
\end{figure*}

Conversely, Convolutional Neural Networks do not exploit human intuitions but only rely on large training datasets and a training procedure based on backpropagation, coupled with an optimization algorithm such as gradient descent. The training procedure aims at automatically learning both the weights of the filters, so that they are able to extract visual concepts from the raw image content, and a suitable classifier. The first layers of the network typically identify low-level concepts such as edges and details, whereas the final layers are able to combine low-level features so as to identify complex visual concepts. Convolutional Neural Networks are typically trained resorting to a supervised procedure that, besides learning ad-hoc features, defines a classifier as the last layer of the network, as shown in Figure~\ref{fig:CNN}. Despite being powerful and effective, the interpretability of such models is limited. Moreover, being very complex model consisting of up to hundreds of millions of parameters, CNNs need large annotated training datasets to yield accurate results. 

In the context of pedestrian detection, the last layer typically consists of just one neuron, and acts as a binary classifier that determines whether an input region depicts a pedestrian. The higher the output of such neuron, the higher the probability of the corresponding region containing a pedestrian. Binary classification is obtained by properly thresholding the output score of such neuron.


\section{Optimizing deep convolutional networks for pedestrian detection}
\label{sec:method}

The use of Convolutional Neural Networks in the context of pedestrian detection is recent, and the potential of such approach is still unexplored. In the following we will present our pipeline, thoroughly illustrating all its stages. In the context of deep learning, often small details are key to reaching good results in terms of accuracy. By carefully analyzing and optimizing each step of the pipeline, we significantly improve the performance of traditional methods based on handcrafted features.

\subsection{Region proposals}\label{sec:regions}

As introduced in Section~\ref{sec:background}, the first stage of the detection pipeline consists in identifying candidate regions that could possibly depict pedestrian. This stage is key to both computational efficiency and task accuracy. On the one hand, by efficiently discarding most of the negative regions, the number of windows to be fed to the following stages can be reduced by up to three order of magnitude. This is of paramount importance when feature extraction is demanded to a computational-intensive Convolutional Neural Network. On the other hand, the algorithm should not discard many positive regions, as this would severely affect the overall accuracy of the system. 

We tested three different strategies for this stage:
\begin{itemize}
	\item \emph{Sliding window}, the most naive algorithm for proposing candidate regions. According to such approach, the frame is scanned both horizontally and vertically with a window that is shifted by a given stride. To be invariant to the size of pedestrians, regions at different scales can be extracted. On the one hand, such algorithm guarantees a 100 percent recall, since it does not filter out any positive region. On the other hand, it yields a very large number of regions to be fed to the following stages, dramatically increasing the computational burden.
	\item \emph{Selective Search~\cite{Uijlings13}}, a general-purpose algorithm that propose class-independent regions that could possibly contain objects. Such algorithm has been successfully exploited in the context of object detection~\cite{RCNN}, in conjunction with a CNN for feature extraction and region classification. It acts as a coarse filter, significantly reducing the number of regions to be processed by the feature extractor and thus reducing the computational burden.
	\item \emph{Locally Decorrelated Channel Features (LDCF)~\cite{LDCF}}, an ad-hoc pedestrian detection algorithm. Even though such algorithm is able to detect pedestrian with a good precision, we would like to further improve the performance of the system, operating LDCF as a region proposal algorithm coupled with a neural network. In particular, the output of LDCF consists of a possibly large set of regions, each with a confidence value. The higher the confidence score of a region, the more likely such region contains a pedestrian. Setting a threshold on the confidence score allows for a tradeoff between precision and recall.
\end{itemize}

\subsection{Fine tuning for pedestrian detection}\label{sec:finetuning}

Learning a deep convolutional network consisting of up to hundreds of millions of parameters requires massive annotated training datasets. In the context of object detection, such models are usually trained resorting to the ImageNet~\cite{imagenet_cvpr09} dataset, composed of 1.2M images annotated with the bounding boxes corresponding to the objects. In particular, the ground truth labels discriminate between 1k different object classes. In the context of pedestrian detection, annotated training dataset having that dimensions are not publicly available. Nonetheless, the complex models trained on ImageNet proven to be a good starting point for accomplishing tasks different from object classification~\cite{karayev2013recognizing}. In fact, the features that are extracted by the first layers of an architecture trained on ImageNet capture simple yet important visual concepts, that are quite general and can be adapted to other kind of tasks. In this case, a \emph{finetuning} process is often employed: starting from the general-purpose neural network, few epochs of training on the target dataset with a small learning step are usually performed so as to adapt the convolutional network to the new task. 

We start from different convolutional neural networks trained on ImageNet and successfully employed for object detection. Then, we exploit an annotated training dataset of positive and negative regions, i.e. regions containing a pedestrian or other kind of visual content, respectively, to finetune the weights of the convolutional network and the classifier.

\subsection{Data preprocessing and augmentation}

Some data preprocessing during the training procedure can be useful to improve both the accuracy and the robustness of the system. In particular, such preprocessing is applied to the regions provided by the first stage of the pipeline, i.e. the region proposal algorithm.

\begin{figure}[t]
	\centering
    \subfigure[]
	   {\includegraphics[trim=1cm 1cm 1cm 1cm, clip=true, width=0.17\textwidth]{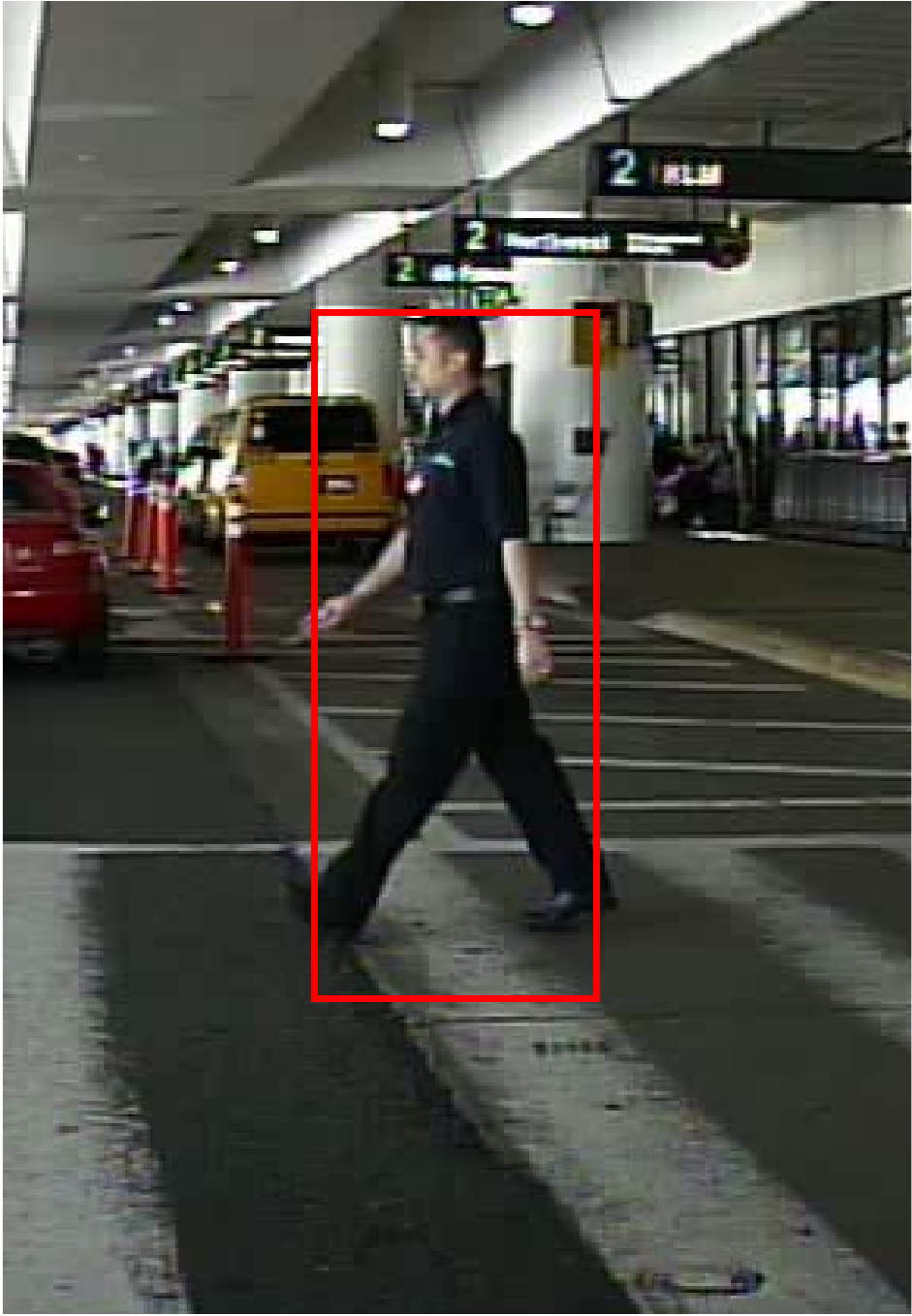}}
	 \hspace{5mm}
	 \subfigure[]
	   {\includegraphics[trim=1cm 1cm 1cm 1.7cm, clip=true, width=0.17\textwidth]{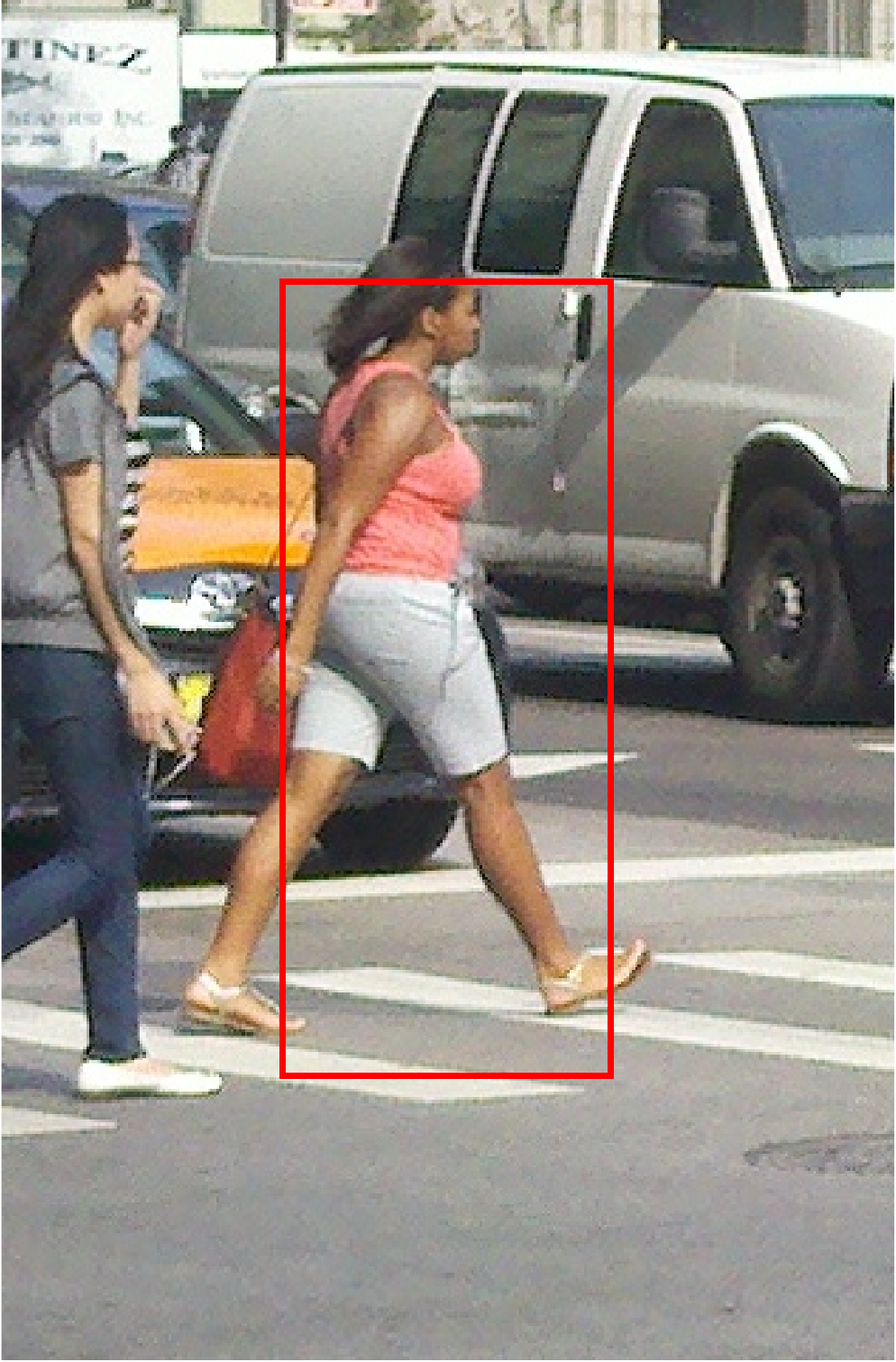}}
	\caption{(a) a region proposal that precisely defines the contour of the human body; (b) an imprecise region proposal that cuts part of the body.}
	\label{fig:bb}
\end{figure}

\textbf{\emph{Padding: }}in the training dataset, the bounding boxes corresponding to pedestrians precisely delimit the human body. This is often not the case for region proposals, as shown in Figure~\ref{fig:bb}. In fact, the bounding boxes provided by the region proposal algorithm are often imprecise and fail at correctly delimiting the human body. To overcome this issue, we decided to pad the regions provided by the region proposals algorithm. To decide the amount of padding to be applied to each region, we employed the following procedure:

\begin{enumerate}
	\item run the region proposal algorithm so as to extract the candidate bounding boxes;
	\item for each bounding box belonging to the training set and provided by the ground truth annotations, identify the closer candidate region provided by the proposal algorithm;
	\item measure the amount of padding that is necessary so that the ground truth bounding box is fully contained by the candidate region.
	\item build the histogram of the padding quantities. Since the histogram represents a nearly-gaussian distribution, the mean value is used as reference padding value from now on.
\end{enumerate} 

To further improve the robustness of the model, multiple random crops of the padded region are actually fed to the convolutional network for training, so as to simulate the uncertainty of region proposals.

\textbf{\emph{Negative sample decorrelation: }}as mentioned in Section~\ref{sec:finetuning}, the convolutional network training procedure exploits annotated positive and negative image regions. As regards the positive regions, they naturally come from the ground truth bounding boxes corresponding to pedestrians. As for the negative samples, the training datasets do not usually provide them directly, but they can be sampled from the visual content following a number of different criteria. We propose a greedy algorithm based on color histograms~\cite{Stockman:2001:CV:558008} to select a set of negative training examples that are as diverse as possible. Let $\mathcal{I} \in [0, 255]^{M \times N \times P}$ denote the image patch corresponding to a sample region, where $M$, $N$ and $P$ denote the number of rows and columns and the depth of the image (e.g., $P = 3$ for an RGB image), respectively. Scalar quantization is performed on the value of the pixel intensity in all the channels, with a quantization step size equal to $\Delta$, i.e.

\begin{multline}
	\mathcal{I}_q(m,n,p) = \floor{\mathcal{I}(m,n,p) / \Delta}, \\ m = 1, \dots, M, n = 1, \dots, N, p = 1, \dots, P.
\end{multline}

\begin{figure}[t]
	\centering
    \subfigure[]
	   {\includegraphics[trim=0cm 0cm 0cm 0cm, clip=true, width=0.22\textwidth]{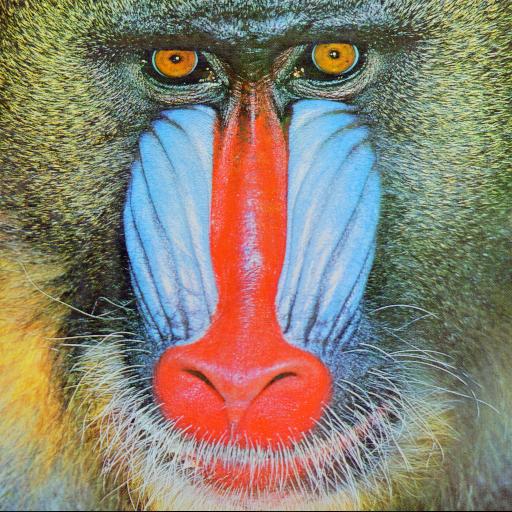}}
	 \hspace{5mm}
	 \subfigure[]
	   {\includegraphics[trim=0cm 0cm 0cm 0cm, clip=true, width=0.22\textwidth]{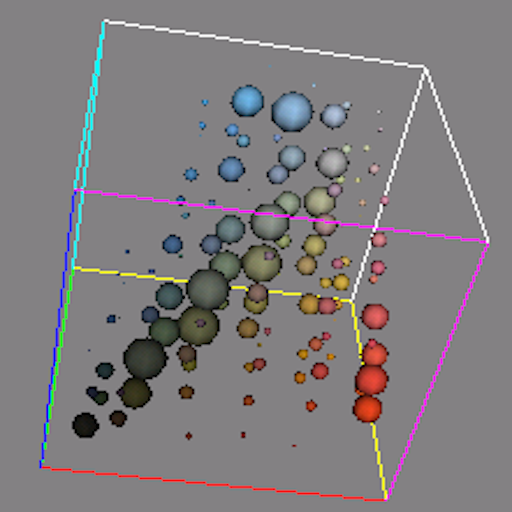}}
	\caption{\textbf (a) an image and (b) its RGB color histogram. The radius of a sphere indicates the number of occurrences of the corresponding quantized color.}
	\label{fig:ch}
\end{figure}

The result of such operation is a quantized color vector $\mathcal{I}_q(m, n) \in \mathbb{R}^P$ for each pixel location $(m, n)$. In the case of RGB images, $\mathcal{I}_q(m, n)$ is a vector with three values, corresponding to the RGB channels. A histogram of the quantized color vectors is then computed. In particular, such histogram counts the occurrences of any possible quantized color vector within the quantized image. Figure~\ref{fig:ch} shows a visual representation of an RGB color histogram. The histogram is then normalized resorting to L2-norm, so that the normalized histogram approximates the probability distribution of quantized color vector in a region.

Then, defining K as the target number of negative regions that have to be extracted, the negative region selection algorithm comprises the following steps:

\begin{enumerate}
	\item starting from the training dataset, randomly extract $N$ regions that do not contain a pedestrian, i.e. that do not overlap with the ground truth bounding boxes;
	\item compute the color histogram of all the negative training regions, and normalize it so as to obtain a color probability distribution;
	\item compute the euclidean distance between the cumulative color probability distributions of each pair of negative regions; 
	\item select the negative region with the highest average distance with respect to all the other regions and remove it from the pool; \label{asd}
	\item in to step~\ref{asd} until K regions are selected.
\end{enumerate}

%

\subsection{Region proposal scores}\label{sec:scores}

As presented in Section~\ref{sec:regions}, a region proposal algorithm must be employed to provide the neural network with a set of candidate windows. Typically, the region proposal algorithm yields a set of regions, each with a confidence value that indicates the probability of the given region containing a pedestrian. Such scores can be used to select which regions have to be fed to the neural network for classification. In this case, only regions with a score higher than a given threshold are fed to the following stages, so that in the final detection pipeline the region proposal algorithm and the neural network are used in series.

Alternatively, besides acting as a selector for region proposal, the score of a region can be exploited as an additional feature for the final classifier that assesses the presence of a pedestrian. In a sense, the scores provided by the region proposal algorithm and by the neural network are used in parallel to classify a given region, improving the system accuracy.

\section{Experimental evaluation and results}
\label{sec:experiments}

We perform a thorough experimental campaign to assess the effectiveness of our pedestrian detection pipeline, highlighting how the optimizations introduced in Section~\ref{sec:method} improve the overall performance.  

\subsection{Datasets}

To conduct our experiments, we resort to the \emph{Caltech Pedestrian Dataset}~\cite{dollar2012pedestrian}, a widely accepted, large-scale, challenging dataset that has become the standard to evaluate the results of pedestrian detection algorithms~\cite{Benenson2014Eccvw}. It consists of approximately 10 hours of video content, acquired from a vehicle driving through regular traffic in an urban environment under good weather conditions. The resolution and the sampling rate of the video content are $640\times480$ and 30fps, respectively.

Approximately 250k frames, corresponding to 137 minutes of video content, have been manually annotated with approximately 350k bounding boxes, corresponding to 2300 different pedestrians. The annotations include temporal correspondence of the same pedestrian across different frames and information about occlusions, such as the visible area and the full extent of the pedestrian. Approximately half of the frames do not contain any pedestrian, whereas 30\% of the frames contain two or more. On average, a pedestrian is visible for about 5 seconds. 

Furthermore, an evaluation protocol has been released along with the dataset, so that the performance of the different algorithms can be directly compared and analyzed. For the sake of completeness, we briefly report here the evaluation protocol. The \emph{Caltech Pedestrian Dataset} was captured over 11 sessions. The first six sessions, i.e. Session 0 to Session 5, are assigned to the training set, whereas the other sessions, i.e. Session 6 to Session 10, are assigned to the test set. To avoid using very correlated frames, the test set is resampled so that 1 image every 30 frames is considered for evaluation. As regards training, we resampled the video sequence so that 1 frame out of 3 is used, as suggested by Hosang et al.~\cite{Hosang2015Cvpr}. Table~\ref{tab:caltech} reports some statistics on both the training and the test data.

\begin{table}[]
\centering
\caption{Statistics for the \emph{Caltech Pedestrian Dataset}}
\label{my-label}
\begin{tabular}{|l|l|l|l|}
\hline

{set}                                            & {session} & {\# images} & {\# positive regions} \\ \hline \hline

                                            & {0}       & {8559}      & {7232}                \\
 
                                            & 1                              & 3619                             & 2903                                       \\

                                            & {2}       & {7410}      & {588}                 \\

                                            & {3}       & {7976}      & {3023}                \\

                                            & 4                              & 7328                             & 1235                                       \\

\multirow{-6}{*}{train}                       & {5}       & {7890}      & {1394}                \\ \hline

                      & 6                              & 1155                             & 903                                        \\

                  & 7                              & 746                              & 1297                                       \\

                   & 8                              & 657                              & 352                                        \\

                     & 9                              & 738                              & 557                                        \\

\multirow{-5}{*}{{test}} & 10                             & 728                              & 776                                        \\ \hline
\end{tabular}
\label{tab:caltech}
\end{table}

\subsection{Evaluation Metrics}

We resort to the evaluation metrics defined by the \emph{Caltech Pedestrian Detection} evaluation protocol, as proposed by Doll\'ar et al.~\cite{dollar2012pedestrian}. In particular, the performance of an algorithm is evaluated in terms of the tradeoff between the \emph{miss rate} (MR) and the number of \emph{false positives per image} (FPPI). For the convenience of the reader, we briefly recap such metrics. First, a detected bounding box, i.e. the one provided by a pedestrian detection algorithm, and a ground truth bounding box are considered to match if the area covered by their intersection is greater than 50\% of the area of their union. A ground truth bounding box that does not have a match is considered a False Negative (FN), or a \emph{miss}. A detected bounding box that does not have a matching ground truth bounding box is considered as a False Positive (FP). Then, it is straightforward to define the average number of \emph{false positives per image} (FPPI), that is, the average number of regions of each image that are erroneously detected as a pedestrian. The \emph{miss rate} (MR) is defined as the ratio between the number of False Negatives and the total number $P$ of positive examples (the number of ground truth bounding boxes), i.e.

\begin{equation}
	MR = \frac{FN}{P}.
\end{equation}

We are particularly interested in the value of the \emph{miss rate} at 0.1 FPPI, that has been identified as a reasonable working condition for a real-world system.

\subsection{Experimental setup and results} 

We tested two state-of-the-art convolutional neural networks that proved to be very effective in the context of object detection: 
\begin{itemize}
	\item \emph{AlexNet}~\cite{krizhevsky2012imagenet}, in particular to its version finetuned for general-purpose detection, as proposed by Girshick et al.~\cite{RCNN};
	\item \emph{GoogLeNet}~\cite{GoogLeNet}, the winner of 2014 ImageNet object detection challenge.
\end{itemize}
	Such neural networks are able to classify regions belonging to 200 different classes, most corresponding to different objects or animals. As a first experiment, we use each network as is, and we consider the class \texttt{person} as the one corresponding to pedestrians. In particular, given an input region provided by the region proposal algorithm, we run the neural network and we obtain the output class label. If such a label is \texttt{person}, we assume that a pedestrian has been detected. 

\textbf{\emph{Region Proposals: }}we resort to three increasingly specific methods for proposing candidate regions, as introduced in Section~\ref{sec:method}. The most naive method is based on multi-scale sliding windows that scan the entire image. We perform a simple test to get some information from the \emph{Caltech Pedestrian Dataset}. First, the evaluation protocol suggested by Doll\'ar et al.~\cite{dollar2012pedestrian} operates under reasonable conditions, that is, with pedestrians whose height in pixels is greater than 50, under no occlusions. Hence, we fix the minimum vertical window size to 50 pixels. Analyzing the average dimensions of the ground truth bounding boxes for the dataset at hand, it is possible to see that most annotations refer to pedestrian whose height in pixels is less than 100, and that the average aspect ratio between the height and the width of the bounding boxes is approximately 2.4:1. 

The maximum height and the aspect ratio for a bounding box are thus set to 100 pixels and 2:1, respectively. Table~\ref{tab:swstat} reports the main parameters that we set for the sliding window approach.

\begin{table}[]
\centering
\caption{Parameters for the sliding window region proposal}
\begin{tabular}{|l|l|}
\hline

Aspect ratio                  & 2:1                              \\

Minimum size                  & $50 \times 25$                   \\

Maximum size                  & $200 \times 100$                 \\

Scale step                    & 1.1                              \\

{Stride} & {10 pixels} \\ \hline
\end{tabular}
\label{tab:swstat}
\end{table}

The second method that we tested is \emph{Selective Search}~\cite{Uijlings13}, a general-purpose region proposal algorithm that achieve outstanding results in conjunction with AlexNet for object detection. We use the default parameters suggested by Girshick et al.~\cite{RCNN}.

As mentioned in Section~\ref{sec:method}, we also tested a region proposal algorithm tailored to pedestrian detection, namely Locally Decorrelated Channel Features (LDCF). Such algorithm acts as a coarse filter, discarding as many negative regions as possible. 
All the regions extracted with such setting are then fed to the neural network for feature extraction and classification.

Convolutional Neural Networks accept input patches with fixed dimensions. In particular, AlexNet requires the input regions to be squared, with a resolution of $227 \times 227$ pixels. The regions extracted by the region proposal algorithms are thus resized to comply with such requirement, as commonly done for object detection~\cite{RCNN}. Table~\ref{tab:proposals} compares the performance of the different region proposal algorithms, including two traditional algorithms such as HOG~\cite{Dalal} and LDCF~\cite{LDCF}, set with the default threshold, i.e. the one that yield best performance on the dataset at hand, as a reference. \emph{Selective Search} is not able to propose all the regions corresponding to pedestrians. In particular, a careful analysis reveal that it fails at proposing almost 50\% of the positive bounding boxes. That is, even if classification is performed perfectly, the miss rate can not be lower than 0.5. \emph{Selective Search} is thus clearly not suitable for such specific task. On the other hand, the sliding window approach significantly improves the detection accuracy. In fact, such approach presents the higher recall, by proposing a very large number of regions. In this case, the neural network is to be blamed, not being able to correctly classify that many regions. Finally, LDCF seems to offer good performance as a region proposal algorithm. Nonetheless, the Convolutional Neural Network does not seem to be doing a good job detecting pedestrians.

\begin{table}[]
\centering
\caption{Comparison of different region proposal algorithms coupled with AlexNet-Pedestrian, finetuned on the Caltech Pedestrian dataset, and the original AlexNet, trained for general-purpose detection, in terms of miss rate at 0.1 False Positives per image.}
\label{tab:proposals}
\begin{tabular}{|ll|l|}
\hline
 
Region proposals           & Feature extraction          & $\text{miss rate}_{\text{0.1FPPI}}$ \\ \hline \hline

Selective search           & AlexNet-Pedestrian                    & 0.801    \\

Sliding window             & AlexNet-Pedestrian                    & 0.370    \\

LDCF                       & AlexNet-Pedestrian                    & 0.197    \\
\hline

Selective search           & AlexNet                     & 0.820    \\

Sliding window             & AlexNet                     & 0.616    \\

LDCF                       & AlexNet                     & 0.398    \\
\hline

\multicolumn{2}{|l|}{Viola-Jones} & 0.950    \\

\multicolumn{2}{|l|}{HOG}         & 0.680    \\

\multicolumn{2}{|l|}{LDCF}        & 0.248    \\ \hline
\end{tabular}
\end{table}

\begin{figure}[t]
	\centering
    \includegraphics[trim=0cm 0cm 0cm 0cm, clip=true, width=0.49\textwidth]{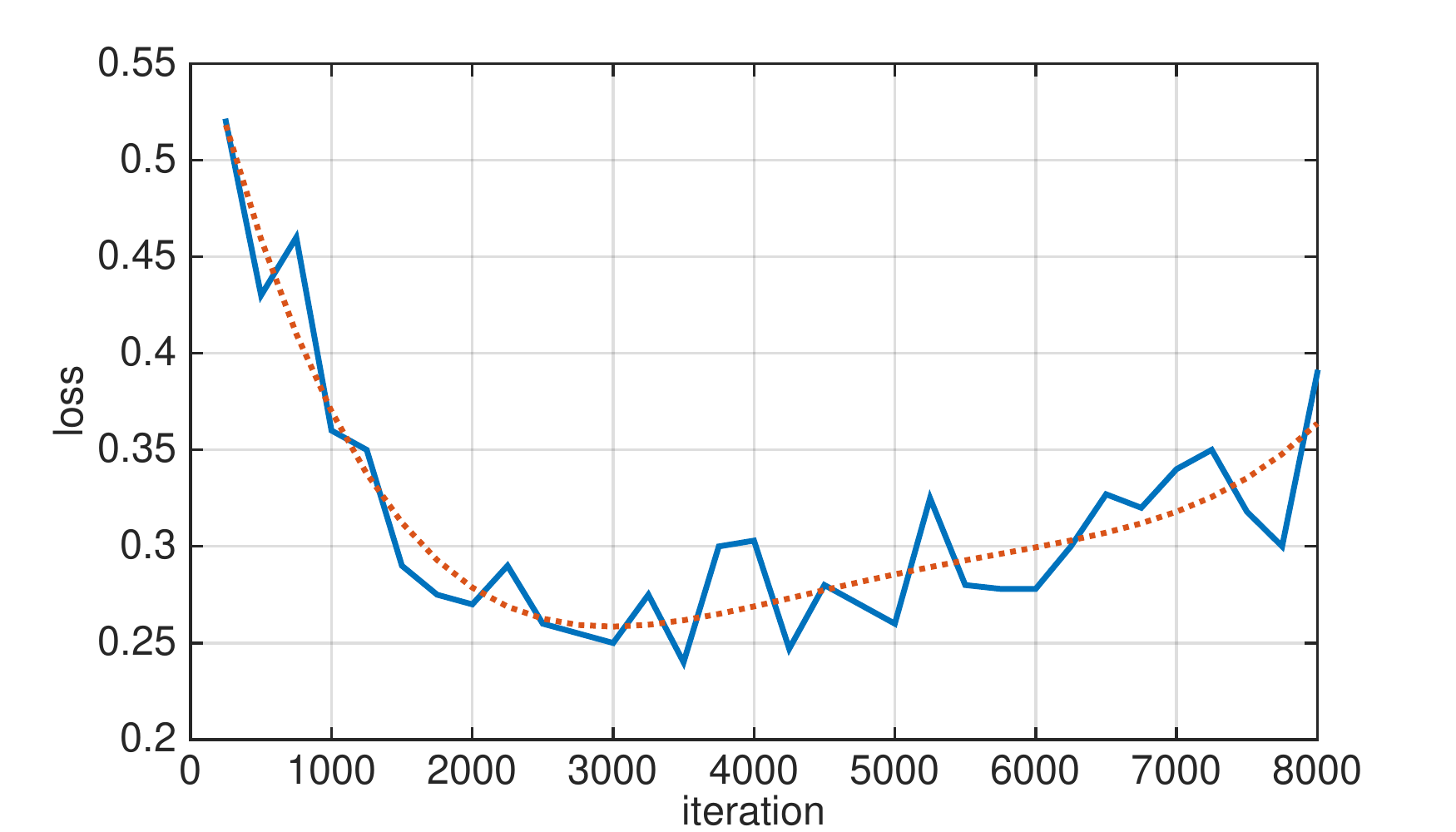}
	\caption{The loss function on the validation subset as a function of the number of iterations. The continuous line represents the average over the 6-folds. The dotted line is a smoothed version of the continuous line, used to determine the minimum loss as function of the number of iterations.}
	\label{fig:loss}
\end{figure}

\textbf{\emph{Data preprocessing and finetuning: }}the preliminary tests show that the neural network is not doing a good job detecting regions depicting pedestrians. This is not surprising, since we are using a network trained for object detection, without any modification. To improve the performance, we resort to a finetuning procedure that aims at adapting the network to the specific task of pedestrian detection. We perform a k-fold cross validation procedure with $k = 6$ folds. In particular, the training dataset consists of 6 different sessions: for each fold, one of them is used as a validation set, whereas the remaining five constitute the training set.
We use a negative to positive region ratio equal to 5:1, as suggested by Hosang et al.~\cite{Hosang2015Cvpr}. Figure~\ref{fig:loss} shows the average loss function obtained by finetuning the neural network over the 6 folds. We identify iteration 3000 as a good candidate to stop the training procedure, being at the beginning of the region in which the loss function stops decreasing. To exploit the entire training dataset, we finetune the model resorting to the entire training dataset and stopping at iteration 3000. Furthermore, we substitute the classifier learned during the finetuning procedure, based on the softmax function and log loss, with a linear SVM, trained with the regularization parameter $C$ set to $10^{-2}$.

\begin{figure}[t]
	\centering
    \includegraphics[trim=1.5cm .5cm .8cm .8cm, clip=true, width=0.49\textwidth]{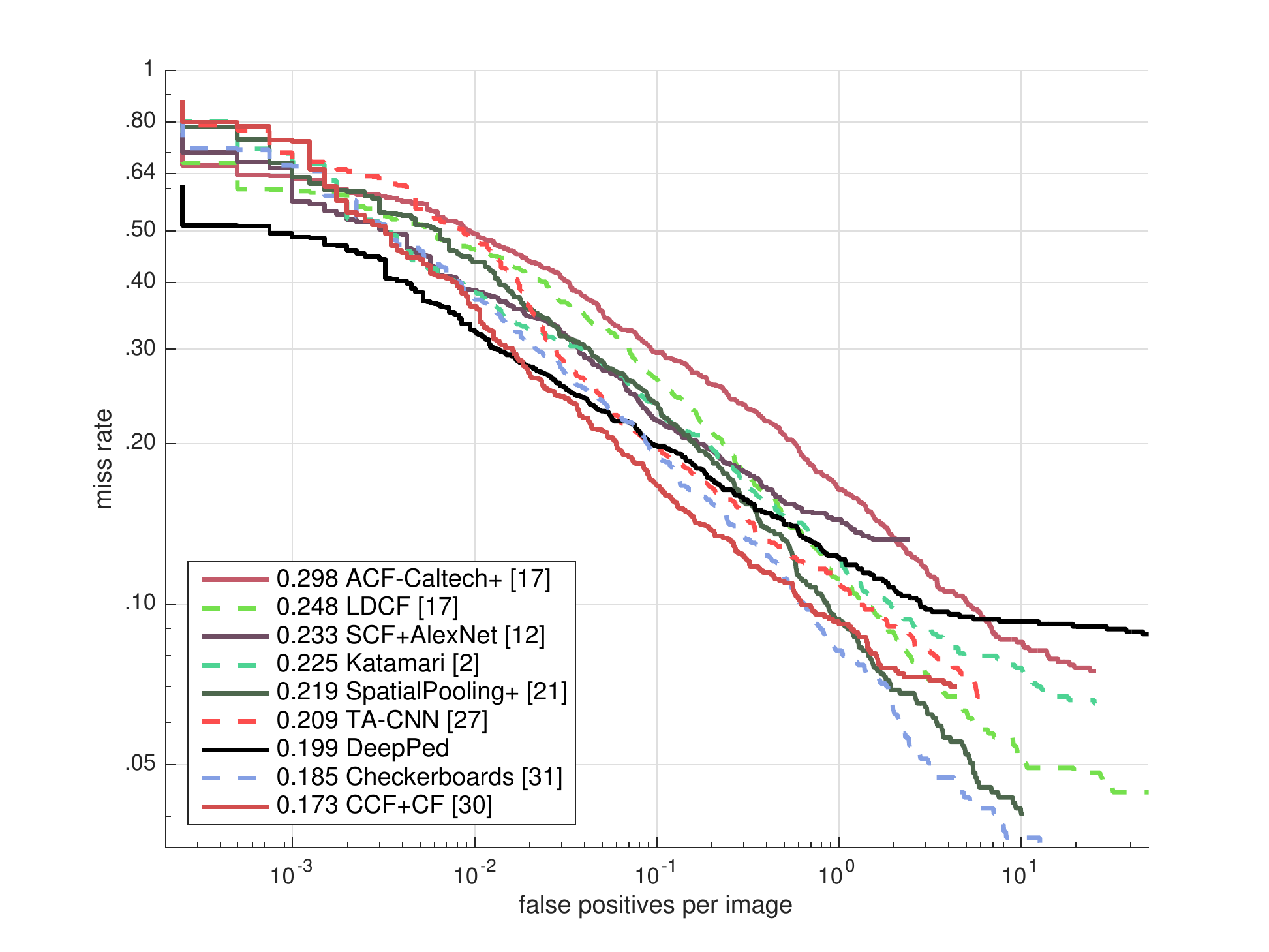}
	\caption{\textbf{Best viewed in colors.} Comparison between \texttt{DeepPed} and other popular pedestrian detection algorithms}
	\label{fig:final}
\end{figure}

\textbf{\emph{Final model: }}our final system is a combination of LDCF as region proposal algorithm and the finetuned deep convolutional neural network---either AlexNet-Pedestrian or GoogLeNet-Pedestrian---and comprises all the optimizations presented in the previous paragraphs. With respect to \texttt{SCF+AlextNet} by Hosang et al.~\cite{Hosang2015Cvpr} we deeply evaluated how to choose positive and negative regions for training. Moreover we made use of the score produced by the LDCF detector in the second level SVM, so to preserve all the information extracted from the image. We label the whole pipeline \texttt{DeepPed}, and we evaluate its performance on the entire test set. The two neural networks perform similarly in terms of task accuracy: in particular, AlexNet-Pedestrian and GoogLeNet-Pedestrian yield a miss rate equal to \textbf{0.199} and \textbf{0.197}, respectively. Due to the higher complexity of GoogLeNet, we decided to show and consider only AlexNet-Pedestrian as feasible CNN for our \texttt{DeepPed} pipeline.
Figure~\ref{fig:final} compares the performance of our system with that of other popular algorithms. Channel Features based algorithms as \texttt{LDCF} and \texttt{ACF-Caltech+}~\cite{LDCF} - a deeper and more sophisticated version of \texttt{ACF} - are easily beaten by a large margin. Our reference starting-point architecture \texttt{SCF+AlexNet} by Hosang et al.~\cite{Hosang2015Cvpr} and the complex \texttt{Katamari}~\cite{Benenson2014Eccvw} proposal, composed of several previous developed handcrafted methods,  are performing worse the our approach by quite a big margin.
Complex methods built upon hundred of feature channels and resorting to optical flow and semantic information, such as \texttt{SpatialPooling+}~\cite{paisitkriangkrai2014pedestrian} and \texttt{TA-CNN}~\cite{tian2014pedestrian}, are also beaten by a small margin by \texttt{DeepPed}.
Newly and sophisticated methods as \texttt{Checkerboards}~\cite{zhang2015filtered} and \texttt{CCF+CF}~\cite{yang2015convolutional}, based on low-level features extracted from pretrained CNNs and boosted classifiers, are instead better than \texttt{DeepPed} in terms of miss rate, even if their complexity is much higher.

\textbf{\emph{Analysis of computational time and optimization: }}we profiled the execution of our system on a desktop architecture which features a six-core 2.4GHz Intel Xeon CPU E5-2609, a NVIDIA GTX980 GPU and 32GB of RAM. The systems requires, on average, 530ms to process a frame at a resolution of $640 \times 480$ pixels. As a term of comparison, \texttt{CCF} requires more than 3 seconds to process a single frame~\cite{yang2015convolutional}.

 In our setting, LDCF is run on the CPU resorting to the original implementation provided by the authors~\cite{LDCF}, without any other optimization, whereas the convolutional neural network is run on the GPU resorting to the Caffe framework~\cite{jia2014caffe}. Note that LDCF, operated as region proposal algorithm, accounts for most of the computational time. In particular, LDCF and AlexNet-Pedestrian accounts for approximately 511ms and 19ms, respectively. We thus feel that a careful optimization of the Matlab-based LDCF implementation provided by authors~\cite{LDCF} may significantly improve the computational performance of our system.

\textbf{\emph{Lightweight version: }}to improve the efficiency of our algorithm, we resort to a more efficient algorithm for region proposal, namely ACF~\cite{ACF}. We carefully analyzed the original code provided by authors~\cite{ACF} and rewrite an optimized version of the algorithm in C. Such implementation yields a 0.298 miss rate - the same as the original Matlab code - and can be operated in real-time on the same machine, requiring as few as 46ms per frame, corresponding to a processing rate of \textbf{21.7fps}. Such solution reduces the computational time by more than 10 times. To further validate our approach, we tested such configuration on an NVIDIA Jetson TK1, a development platform equipped with a 192-core NVIDIA Kepler GPU, an NVIDIA quad-core ARM Cortex-A15 CPU and 2GB or memory. Our system based on ACF and AlexNet-Pedestrian requires 405ms per frame, corresponding to 2.4fps on the development board.

\section{Conclusions}\label{sec:conclusions}

We proposed a pedestrian detection system based on convolutional neural networks. The proposed system outperforms alternative approaches based on both handcrafted and learned features, at a reasonable computational complexity. Our lightweight version is capable of detecting pedestrian in real-time on modern hardware. As a proof of concept, we tested our system on a development board that is envisioned to be the computational brain of smart cars. Future work will include the optimization of LDCF for region proposal and the implementation of our pipeline entirely on GPU, so as to avoid expensive memory copies and improve the overall performance, with the aim of combining state-of-the-art accuracy and real-time processing.

%
%

\section{Acknowledgements}\label{sec:ack}

The project GreenEyes acknowledges the financial support of the Future and Emerging Technologies (FET) programme  within the Seventh Framework Programme for Research of the European Commission, under FET-Open grant number: 296676.



\section*{References}

\bibliographystyle{elsarticle-harv} 
\bibliography{main}


%
%
%
%
\end{document}